\documentclass[letterpaper, 10 pt, conference]{ieeeconf}  

\IEEEoverridecommandlockouts                              
\usepackage[utf8]{inputenc}
\usepackage[T1]{fontenc}
\usepackage{graphicx}
\usepackage{subcaption}
\usepackage{makecell}

\usepackage{graphics} 
\usepackage{epsfig} 
\usepackage{amsmath} 
\usepackage{amssymb}  
\usepackage{xcolor}
\usepackage[normalem]{ulem}
\usepackage{booktabs}
\usepackage{multicol}
\usepackage{multirow}
\usepackage{cancel}
\usepackage{cite}
\usepackage{cleveref}
\usepackage{comment}
\usepackage{siunitx}
\let\labelindent\relax
\usepackage[inline]{enumitem}

\usepackage{acro}
\usepackage{bm}
\usepackage{mathtools}
\usepackage{xcolor}

\DeclareAcronym{ASL}{short = ASL, long = Autonomous Systems Lab}
\DeclareAcronym{OMAV}{short = OMAV, long = Omnidirectional Micro Aerial Vehicle}
\DeclareAcronym{MAV}{short = MAV, long = Micro Aerial Vehicle}
\DeclareAcronym{UAV}{short = UAV, long = Unmanned Aerial Vehicle}
\DeclareAcronym{DoF}{short = DoF, long = degrees of freedom}
\DeclareAcronym{PBC}{short = PBC, long = passivity-based control}
\DeclareAcronym{PH}{short = PH, long = Port-Hamiltonian}
\DeclareAcronym{NDT}{short = NDT, long = non-destructive testing}
\DeclareAcronym{PEMS}{short = PEMS, long = Power and Energy Monitoring System}
\DeclareAcronym{WTC}{short = WTC, long = wrench tracking controller}
\DeclareAcronym{PTC}{short = PTC, long = pose tracking controller}
\DeclareAcronym{MBE}{short = MBE, long = momentum-based wrench estimator}
\DeclareAcronym{ASIC}{short = ASIC, long = Axis-Selective Impedance Control}
\DeclareAcronym{MPC}{short = MPC, long = Model Predictive Control}
\DeclareAcronym{MPPI}{short = MPPI, long = Model Predictive Path Integral}
\DeclareAcronym{APhI}{short = APhI, long = Aerial Physical Interaction}
\DeclareAcronym{LLE}{short = LLE, long = Largest Lyapunov Exponent}
\DeclareAcronym{ICBF}{short = ICBF, long = Integral Control Barrier Function}
\DeclareAcronym{CBF}{short = CBF, long = Control Barrier Function}
\DeclareAcronym{COM}{short = CoM, long = Center of Mass}
\DeclareAcronym{AM}{short = AM, long = Aerial Manipulator}
\DeclareAcronym{MR}{short = MR, long = Mixed Reality}
\DeclareAcronym{AR}{short = AR, long = Augmented Reality}
\DeclareAcronym{VR}{short = VR, long = Virtual Reality}
\DeclareAcronym{HRI}{short = HRI, long = Human-Robot Interaction}
\DeclareAcronym{RL}{short = RL, long = Reinforcement Learning}
\DeclareAcronym{PPO}{short = PPO, long = Proximal Policy Optimization}
\DeclareAcronym{PETG}{short = PETG, long = Polyethylene Terephthalate Glycol}
\DeclareAcronym{TPU}{short = TPU, long = Thermoplastic Polyurethanes}

\DeclareAcronym{NASA-TLX}{short = NASA-TLX, long = NASA Task Load Index}
\DeclareAcronym{MD}{short = MD, long = mental demand}
\DeclareAcronym{PD}{short = PD, long = physical demand}
\DeclareAcronym{TD}{short = TD, long = temporal demand}
\DeclareAcronym{EF}{short = EF, long = effort}
\DeclareAcronym{PE}{short = PE, long = performance}
\DeclareAcronym{FR}{short = FR, long = frustration}
\DeclareAcronym{SNR}{short = SNR, long = signal-to-noise ratio}
\DeclareAcronym{ANOVA}{short = ANOVA, long = Analyse of Variance}
\DeclareAcronym{FT}{short = F/T, long = force and torque, short-indefinite = an, long-indefinite = a}
\DeclareAcronym{BBT}{short = BBT, long = Box and Block Test}
\DeclareAcronym{ABBT}{short = ABBT, long = Aerial Box and Block Test}
\DeclareAcronym{MOCAP}{short = MOCAP, long = Motion Tracking System}

\renewcommand{\vec}[1]{\bm{#1}}		
\newcommand{\matr}[1]{\bm{#1}}		
\newcommand{\nR}[1]{\mathbb{R}^{#1}}		
\newcommand{\SO}[1]{\mathsf{SO}(#1)}		
\newcommand{\upperRomannumeral}[1]{\uppercase\expandafter{\romannumeral#1}}	

\newcommand{\transpose}{^\top}

\renewcommand{\frame}[1]{\mathcal{F}_{#1}}		

\newcommand{\origin}{O}						
\newcommand{\vX}{\vec{x}}					

\newcommand{\vY}{\vec{y}}					
\newcommand{\vZ}{\vec{z}}					
\newcommand{\pos}{\vec{p}_B}				
\newcommand{\vel}{\vec{v}_B}				
\newcommand{\velRef}{\vec{v}_{B,\text{ref}}}	
\newcommand{\posA}{\vec{p}_A}				
\newcommand{\velA}{\vec{v}_A}				
\newcommand{\acc}{\vec{\dot{v}}_B}				
\newcommand{\accBx}{\vec{\dot{v}}_{B,x}}				
\newcommand{\accBy}{\vec{\dot{v}}_{B,y}}				
\newcommand{\accA}{\dot{\vec{v}}_A^A}				

\newcommand{\rotMat}{\matr{R}}				

\newcommand{\frameW}{\frame{W}}			
\newcommand{\frameB}{\frame{B}}			
\newcommand{\frameA}{\frame{A}}			
\newcommand{\originW}{\origin_W}		
\newcommand{\frameFour}{\frame{4}}			

\newcommand{\originA}{\origin_A}		
\newcommand{\originB}{\origin_B}		
\newcommand{\xW}{\vX_W}				
\newcommand{\yW}{\vY_W}				
\newcommand{\zW}{\vZ_W}				
\newcommand{\xB}{\vX_B}				
\newcommand{\xA}{\vX_A}				
\newcommand{\yFour}{\vY_4}				
\newcommand{\yA}{\vY_A}				
\newcommand{\zA}{\vZ_A}				

\newcommand{\rotMatWB}{\rotMat_B^W}	
\newcommand{\rotMatWBDes}{\rotMat_{B,\text{d}}^W}	
\newcommand{\angVel}{\vec{\omega}_B}
\newcommand{\angAcc}{\vec{\dot{\omega}}_B}

\newcommand{\wrenchGravity}{\bm{g}}

\newcommand{\wrenchCommand}{\wrench_{c}}
\newcommand{\wrenchExt}{\wrench_\text{ext}}
\newcommand{\wrenchEst}{\hat{\wrench}_\text{ext}}

\newcommand{\wrench}{\bm{\tau}}

\newcommand{\posDes}{\vec{p}_{B,d}}

\newcommand{\Coriolis}{\bm{c}}

\title{\LARGE \bf
Design of a Flexible Robot Arm for Safe Aerial Physical Interaction
}

\author{Julien Mellet$^{1}$, Andrea Berra$^{2}$, Achilleas Santi Seisa$^{3}$, Viswa Sankaranarayanan$^{3}$, \\Udayanga G.W.K.N. Gamage$^{4}$, Miguel Ángel Trujillo Soto$^{2}$, Guillermo Heredia$^{5}$, \\George Nikolakopoulos$^{3}$, Vincenzo Lippiello$^{1}$, Fabio Ruggiero$^{1}$
\thanks{The research leading to these results has been supported by the AERO-TRAIN project, European Union's Horizon 2020 research and innovation program under the Marie Skłodowska-Curie grant agreement No 953454. The authors are solely responsible for its content.}
\thanks{$^{1}$PRISMA Lab, Department of Electrical Engineering and Information Technology, University of Naples Federico II Naples, Italy.}
\thanks{$^{2}$CATEC, Advanced Center for Aerospace Technologies, Seville, Spain.}
\thanks{$^{3}$Robotics and AI Team, Department of Computer, Electrical and Space Engineering, Lule\aa University of Technology, Lule\aa, Sweden.}
\thanks{$^{4}$Automation and Control Group, Department of Electrical Engineering and Photonics, Technical University of Denmark, Denmark.}
\thanks{$^{5}$Robotics, Vision, and Control Group
School of Engineering, University of Seville
Seville, Spain.}
\thanks{Corresponding author's e-mail: {\tt\small julien.mellet@unina.it}.}}%

\begin{document}

\maketitle
\thispagestyle{empty}
\pagestyle{empty}

\begin{abstract}
This paper introduces a novel compliant mechanism combining lightweight and energy dissipation for aerial physical interaction. Weighting 400~g at take-off, the mechanism is actuated in the forward body direction, enabling precise position control for force interaction and various other aerial manipulation tasks.
The robotic arm, structured as a closed-loop kinematic chain, employs two deported servomotors. Each joint is actuated with a single tendon for active motion control in compression of the arm at the end-effector. Its elasto-mechanical design reduces weight and provides flexibility, allowing passive-compliant interactions without impacting the motors' integrity. Notably, the arm's damping can be adjusted based on the proposed inner frictional bulges. Experimental applications showcase the aerial system performance in both free-flight and physical interaction. The presented work may open safer applications for \ac{MAV} in real environments subject to perturbations during interaction.
\end{abstract}


\section{INTRODUCTION}

Traditionally, \ac{UAV}s focused on tasks like surveillance and aerial imaging, but the demand for these platforms to engage in physical interactions is rising~\cite{active-interaction, past-future-am, redundant-am}. This demand is especially relevant in situations where human intervention may pose risks or practical challenges, such as inspecting remote or hazardous environments, maintaining tall structures, or handling dangerous substances~\cite{am-lit, survey-am, mr-haptic-nuclear-waste}.

Aerial manipulation has emerged as a burgeoning research subject field, encompassing a variety of robotic manipulators for physical interaction with the environment~\cite{past-future-am}. However, this field faces challenges, notably the need to equip aerial platforms for such tasks while maintaining stability. Common approaches involve rigid manipulator links mounted directly on \ac{UAV}s~\cite{active-interaction, s19061305}, requiring dynamic control to manage interaction forces at the end-effector~\cite{am-lit}.

Motivated by the comparison between soft and rigid manipulation~\cite{trivedi2008soft}, this article aims to surpass conventional rigid robotic systems' limitations. We seek to enhance environmental interaction capabilities, stability, and vibration dampening in \ac{UAV} operations by implementing a flexible, compliant robotic manipulator system~\cite{compliant-mechanism}.

\begin{figure}[tp]
  \centering
  \includegraphics[width=\linewidth]{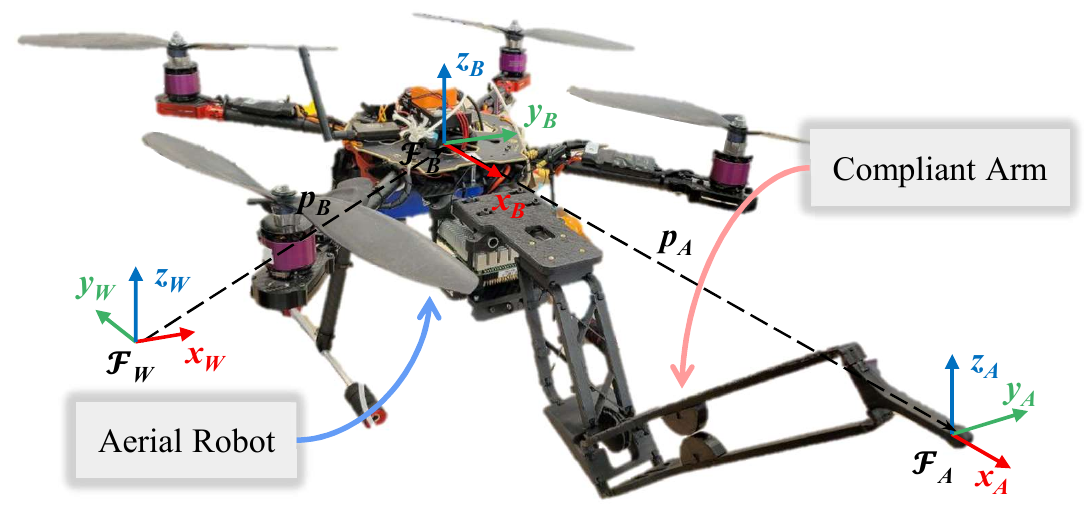}
  \caption{Aerial platform making contact through the proposed compliant robot arm.}
  \label{fig:eye-catcher}
  \vspace{-0.5cm}
\end{figure}

\subsection{Related Work} \label{related_work}
In recent years, compliant robotic mechanisms found applications in various fields. For instance, they were used in bi-manual industrial manipulation~\cite{2017humancompliant}, low-cost cooking applications~\cite{quigley2011low}, and robotic surgery~\cite{sun2016miniature}. However, applying these solutions in aerial manipulation is challenging due to the necessity for lightweight and robust designs. Aerial manipulation faces challenges such as reactive forces that affect safety, limited payload capacity, and the complexity of accurately estimating the position of the end-effector in the global frame~\cite{ladig2021aerial}. These concerns should be addressed when designing an aerial manipulator. Rigid and flexible manipulators offer distinct advantages and drawbacks, with compliance emerging as a critical asset for effective aerial manipulation.

In this concept, a new aerial platform with a soft continuum arm was introduced in~\cite{szasz2022modeling}. Although it showed promising features, its assessment was limited to simulation, revealing oscillation effects. Additionally, its workspace was constrained, and dependent on the drone's extension. Our approach addresses these limitations, enabling the drone to hover while the arm operates, enhancing dexterity.

In other works, the origami delta manipulator was employed for aerial manipulation~\cite{cuniato2023design}, recognized for its robustness and precision. Despite its advantages, this arm presents complexities in its kinematics, cost implications, and a notable drawback in unwanted folding, which can endanger the aerial platform. A similar setup was introduced in~\cite{tzoumanikas2020aerial}, featuring a hybrid model-based algorithm for aerial manipulation. Importantly, this method is versatile and applicable to various types of \ac{MAV}s and manipulators. While these designs are lightweight, they exhibit safety issues.

In~\cite{suarez2021cartesian}, an aerial manipulator composed of a multirotor and a compliant arm with angular and linear deflections is proposed. While it offers advantages in terms of position accuracy and energy consumption, its effectiveness diminishes when the compliant arm is rotated, unlike our approach, which efficiently employs revolute joints. Moreover, their system stores kinetic energy instead of dissipating it. In~\cite{7759655}, the authors introduced a 3-\ac{DoF} robotic arm for aerial manipulation with shoulder pitch and elbow pitch joints. However, it was only tested on a fixed-base bench and lacked validation in an actual aerial manipulation mission.

\subsection{Contributions}
Considering the described gaps, our contributions are 
\begin{enumerate*}[label=\textit{\roman*)}]
    \item the design of a lightweight compliant manipulator that improves robotics \ac{APhI} capabilities,
    \item including safe passive energy dissipation, and
    \item qualitative evaluation of the proposed mechatronic system in real-world aerial conditions.
\end{enumerate*}    

\section{SYSTEM DESCRIPTION}
An ideal robot arm for \ac{APhI} should keep the aerial platform stable and absorb part of the impact at contact. Even if standard interaction strategies try to minimize velocity before contact, an unwanted bounce would remain after impact with the surface. We present both the platform and the contact mechanism of our mechatronic system.

\subsection{Aerial Robot Description} \label{mav_control}

The aerial mission's configuration is established within the inertial world frame, denoted as $\frameW$, which consists of a fixed origin $\originW$ and unit axes ${\xW, \yW, \zW}$ (see Fig.~\ref{fig:eye-catcher}). These axes are positioned such that $\zW$ aligns opposite to the gravitational force. The state of the aerial robot is defined in the body frame, referred to as $\frameB$, which includes the origin $\originB$ located at the \ac{COM} of the robot, and the $\xB$ axis oriented towards the robot's front. To represent the system's spatial state, we utilize the vector $\pos \in \nR{3}$ to express the position of $\originB$ concerning $\frameW$. Furthermore, the orientation of $\frameB$ concerning $\frameW$ is characterized by the rotation matrix $\rotMatWB \in \SO{3}$.
Finally, we denote the body linear and angular velocities as $\vel,\; \angVel \in \nR{3}$, respectively.
With these quantities defined, we can describe the Lagrangian dynamic model as in~\cite{active-interaction}
\begin{equation}
    \label{mav_model}
    \bm{M} \begin{bmatrix} \acc \\ \angAcc \end{bmatrix} + \Coriolis(\angVel) \begin{bmatrix} \vel \\ \angVel \end{bmatrix} + \wrenchGravity(\rotMatWB) = \wrenchCommand + \wrenchExt,
\end{equation}
with body inertia matrix $\bm{M} \in \nR{6\times 6}$, wrench command $\wrenchCommand \in \nR{6}$, external interaction wrench $\wrenchExt\in \nR{6}$, gravity wrench $\wrenchGravity(\rotMatWB) \in \nR{6}$, centrifugal, and Coriolis wrench $\Coriolis(\angVel) \in \nR{6}$.
We adopt the impedance controller in~\cite{active-interaction} adapted to the multirotor dynamic model described in~\cite{aerial-robotics-springer}. This allows us to steer the aerial robot and safely interact with the remote environment with additional impedance compliance. The closed-loop dynamics can thus be described with,
\begin{equation}
    \bm{M}_v \begin{bmatrix} \acc \\ \angAcc \end{bmatrix} + \bm{D}_v \begin{bmatrix} e_v \\ e_{\omega} \end{bmatrix} + \bm{K}_v \begin{bmatrix} e_p \\ e_R \end{bmatrix} = \wrenchEst.
\end{equation}
We denote with $\bm{M}_v \in \nR{6 \time 6}$ the virtual inertia, $\bm{D}_v \in \nR{6 \time 6}$ the damping, and $\bm{K}_v$ the stiffness as parameters of the controller, while $\wrenchEst \in \nR{6}$ is the external wrench applied on the platform. Besides, we define
\begin{subequations}
	\label{position_errors}
	\begin{IEEEeqnarray} {ll}
		\mathbf{e}_p &= \rotMatWB {\transpose} \left( \pos - \posDes \right),\\
        \mathbf{e}_R &= \frac{1}{2}\left( \rotMatWBDes{\transpose} \rotMatWB - \rotMatWB {\transpose} \rotMatWBDes \right)^\vee,\\
        \mathbf{e}_v &= \rotMatWB {\transpose} \left( \vel - \velRef \right),\\
        \mathbf{e}_{\omega} &= \omega_B - \rotMatWB {\transpose} \rotMatWBDes \omega_{B,d},
	\end{IEEEeqnarray}
\end{subequations}
with $\posDes$ and $\rotMatWBDes$ the desired position and orientation, respectively, and $\left(\cdot\right)^\vee$ the Vee operator to extract a vector from a skew-symmetric matrix. 


\subsection{Robot Arm Description}
\label{sec:arm}
Attached to the \ac{UAV}, there is the robot arm with inertial frame corresponding to platform body frame $\frameB$. Its current pose is described by the frame $\frameA=\{\originA,\xA,\yA,\zA\}$ with origin $\originA$ attached at the tip of the end-effector (see Fig.~\ref{fig:eye-catcher}). We denote the position and velocity of the robot arm tip with $\posA \in \nR{3}$ and $\velA  \in \nR{3}$, both expressed with respect to $\frameA$.

Since we assume compliance of the arm, the dynamic relation between the end-effector pose and force value is modeled in $\frameA$ as
\begin{equation}
    \label{arm_model}
    \mathbf{M}_A \accA + \mathbf{D}_A \velA^A + \mathbf{K}_A \posA = -F_A,
\end{equation}
where $F_A \in \nR{3}$ represents the linear interaction force of the end-effector with the environment. The inherent inertia, damping properties, and stiffness of the proposed compliant mechanism are $\mathbf{M}_A \in \nR{3 \times 3}$, $\mathbf{D}_A \in \nR{3 \times 3}$ and $\mathbf{K}_A \in \nR{3 \times 3}$. Section~\ref{sec:design} discusses the parameter values given by the design of the proposed compliant mechanism.

\subsection{Inverse Kinematics of Arm}
To ensure versatility and fault tolerance, we decouple the control of the aerial platform and the control of the robot arm. The multirotor is actuated with the closed-loop low-level attitude controller, while the robot arm is presented in the open-loop form.


\begin{figure}[h]
 \centering
 \includegraphics[width=0.95\linewidth]{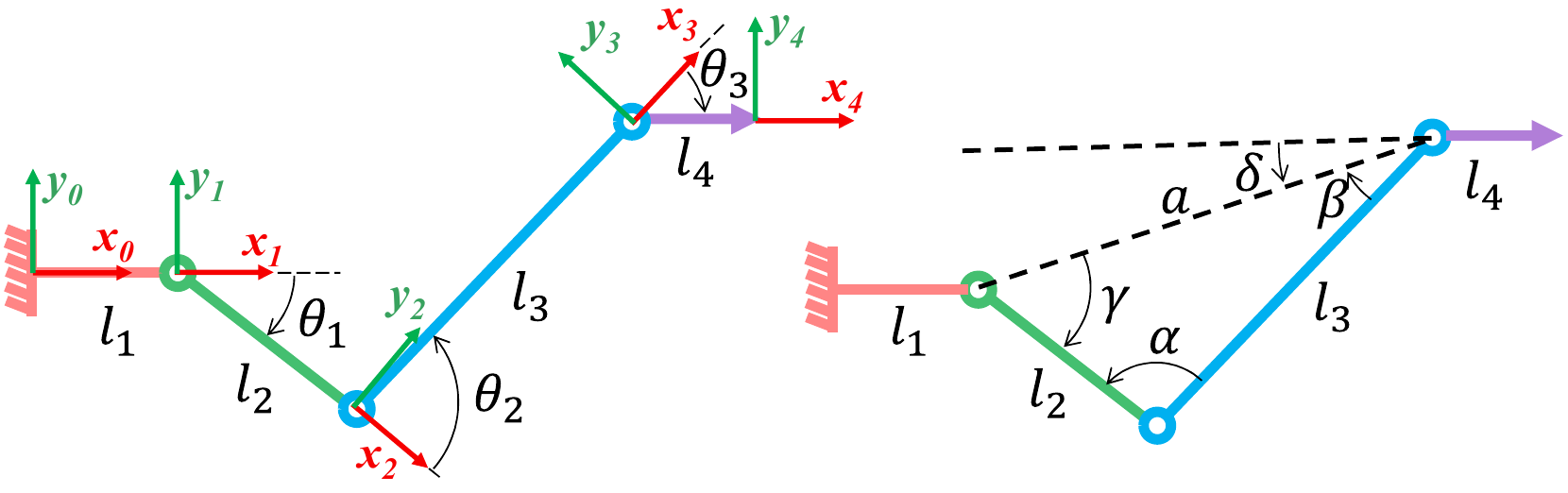}
 \caption{Denavit–Hartenberg parameters of the robot arm for inverse kinematic (left) with transformation from $\frameFour$ to $\frameA$ done by rotation around $x-axis$ where $y_4 = z_A$. Triangle formulation for simplifying the inverse kinematics (right).}
 \label{fig:dh-kinematic}
 \vspace{-0.5cm}
\end{figure}

\begin{figure*}[thpb]
    \subfloat[Kinematic scheme parameters\label{fig:kinematic-design}]{%
        \includegraphics[width=0.32\linewidth]{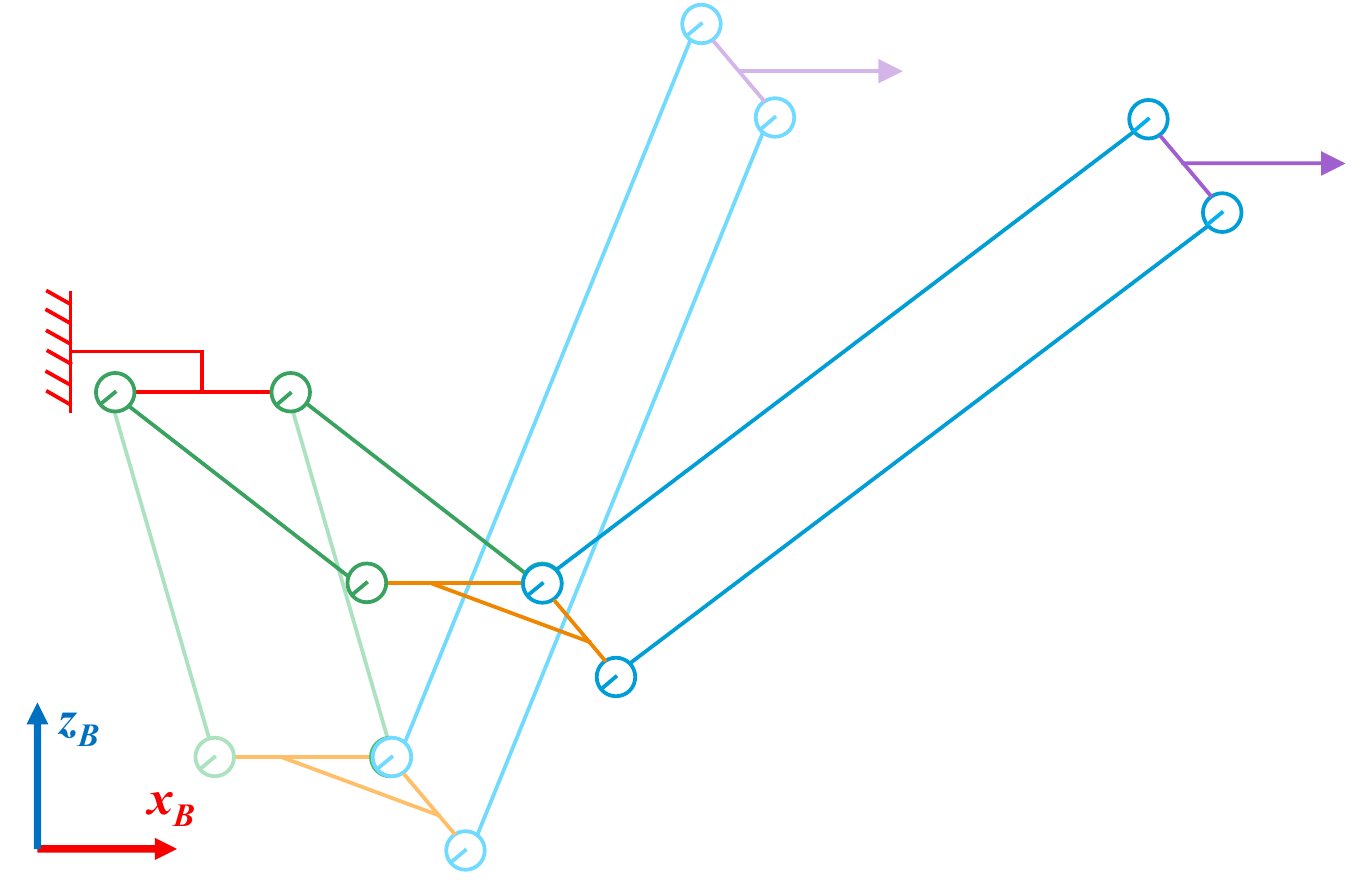}}
    \hfill
    \subfloat[Ideal compliant mechanism\label{fig:compliant-design}]{%
        \includegraphics[width=0.32\linewidth]{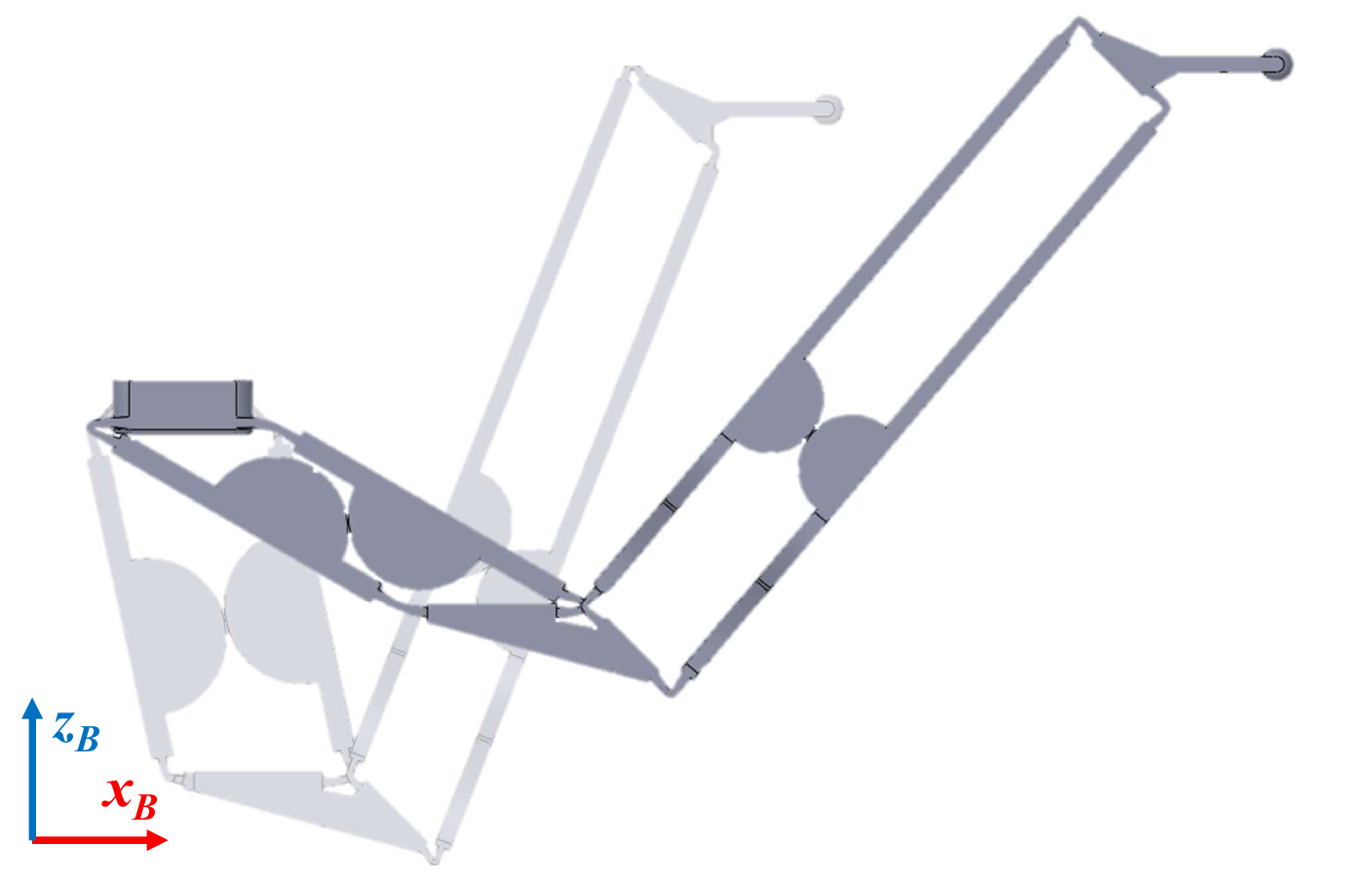}}
    \hfill
    \subfloat[Fine-tuned manufactured arm\label{fig:manufactured-design}]{%
        \includegraphics[width=0.32\linewidth]{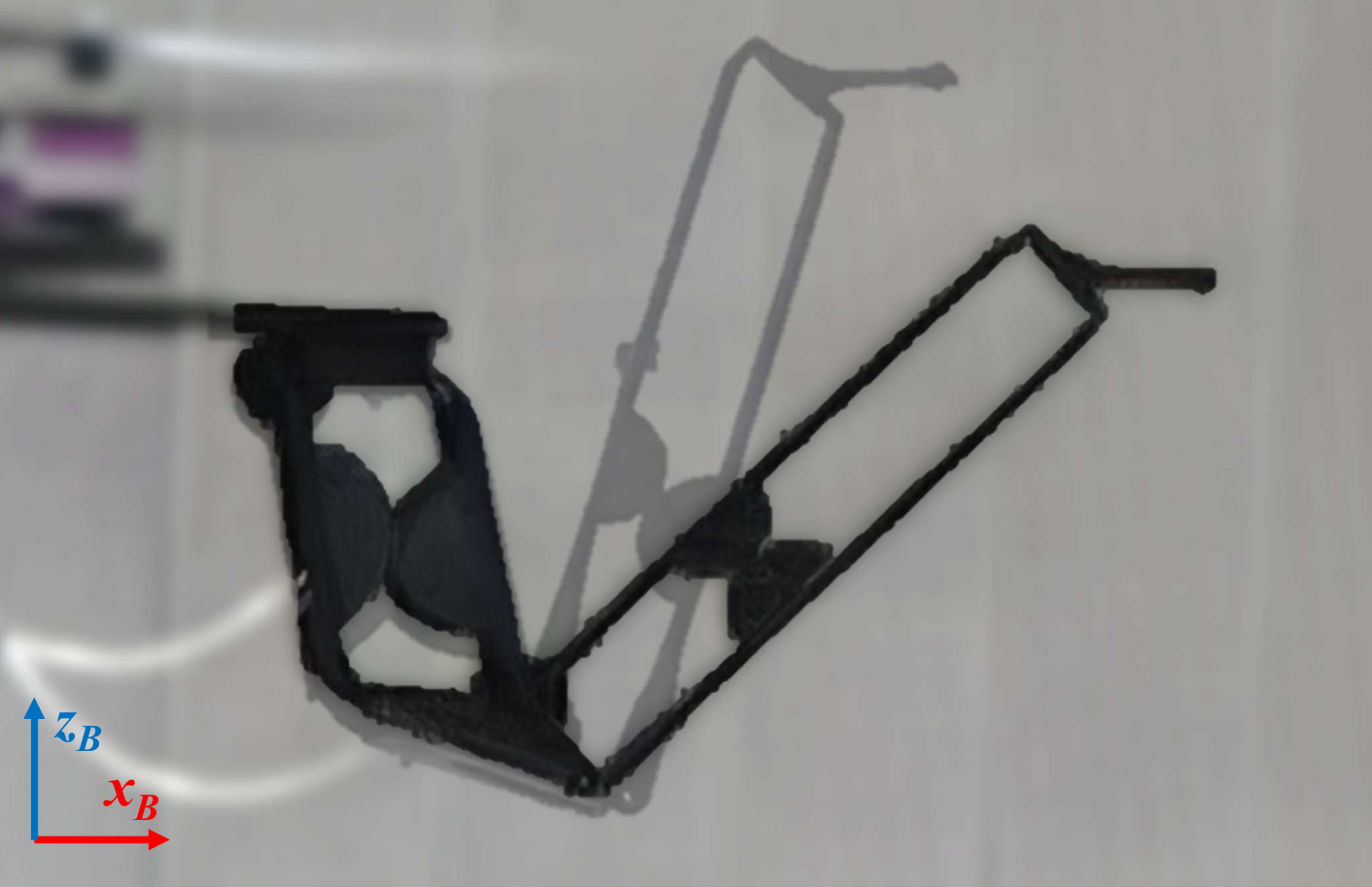}}
  \caption{Comparison of the theoretical kinematics for inverse kinematics, its design into a compliant mechanism, and the real adjusted design. Both idle extension and impact compression (transparent) geometry are presented.}
  \vspace{-0.5cm}
\end{figure*}

The Denavit-Hartenberg convention (Fig.~\ref{fig:dh-kinematic}) involves a transformation from $\frameFour$ to $\frameA$ via rotation around the $x$-axis ($\yFour = \zA$). The end-effector position can be resolved using a geometric method,
\begin{equation}
    p_4^0 = p_1^0 + \bm{R}_1^0 (p_2^1 + \bm{R}_2^1 (p_3^2 + \bm{R}_3^2 p_4^3)),
\end{equation}
where $p_i^j \in \nR{3}$ and $\bm{R}_i^j \in \nR{3 \time 3}$ are the position and rotation in $i$ with respect to $j$, respectively. 
Given the kinematic structure as a series of a four-bar mechanism, the last link is always parallel to the first one,
\begin{equation}
    p_0^3 = p_0^4 - \begin{bmatrix} l_4 & 0 & 0 \end{bmatrix}^\top.
\end{equation}
Since axes $x_1 - y_1 - z_1$ are parallel to $x_4 - y_4 - z_4$ in Fig. \ref{fig:dh-kinematic} (a), the inverse kinematics can be simplified using the relationship in Fig. \ref{fig:dh-kinematic} (b). The angle $\theta_2$ is described by
\begin{equation}
    \theta_2 = -(\pi - \alpha),
\end{equation}
where $\alpha$ is determined using the cosine rule,
\begin{subequations}
	\label{cosine-rule}
	\begin{IEEEeqnarray} {ll}
		a^2 = b^2 + c^2 - 2bc \cos(\alpha),\\
        \alpha = \cos^{-1} \left(\dfrac{b^2 + c^2 - a^2}{2bc}\right),
	\end{IEEEeqnarray}
\end{subequations}
with $b = l_2$, $c = l_3$  and $a = \sqrt{y_3^2 - (x_3 - l_1)^2}$.
Then we can compute $\theta_3$ that is not actuated,
\begin{equation}
    \theta_3 =  \beta + \delta,
\end{equation}
with,
\begin{subequations}
	\begin{IEEEeqnarray} {ll}
        \delta = \tan^{-1}\left(\dfrac{y_3}{x_3}\right) = a \tan^2(-y_3, x_3),\\
        \beta = \cos^{-1}\left(\dfrac{a^2 + c^2 - b^2}{2ac}\right),\\
        a = \sqrt{y_3^2 - (x_3 - l_1)^2}
	\end{IEEEeqnarray}
\end{subequations}

Finally, $\theta_1$ is calculated such that,
\begin{equation}
    \theta_1 = \gamma - \delta,
\end{equation}
\begin{equation}
    \gamma = \cos^{-1}\left(\dfrac{b^2 + a^2 - c^2}{2ab}\right).
\end{equation}

The desired joint angles are then tracked with a PID controlling the servomotors, ensuring precise motion execution as elucidated in the subsequent section.

\section{MECHANICAL DESIGN}\label{sec:design}
This section describes and justifies the design of the proposed robot arm with its manufacturing process.

\subsection{Kinematic Model}
We assume the multirotor near hovering and with forward horizontal velocity along $\xB$. Thus, we propose to use a 2-\ac{DoF}s closed-loop kinematic chain. It consists of stacking in a series of two four-bar parallelograms. This design gives a relatively large workspace and keeps the end-effector horizontal at moving in translation.

As presented in~\cite{compliant-mechanism}, a rigid body mechanism can be transformed into a compliant mechanism by swapping rigid hinges to flexure hinges. In this last case, the hinge is done thanks to section reduction of the beam. Using soft materials, deflection at rotating the joint can be modeled by a rotational spring with an idle position (see Sec.~\ref{sec:absorption-unit}).

Because we want the robot arm to absorb kinetic energy, we want the idle position corresponding to the deployed configuration. Thus, the soft mechanism is in natural extension and goes in flexion at physical impact. We further name the parallelogram as \textit{absorption unit} of our mechanism.

\subsection{Absorption Unit}\label{sec:absorption-unit}

\begin{figure}[h]
    \centering
    \includegraphics[width=.9\linewidth]{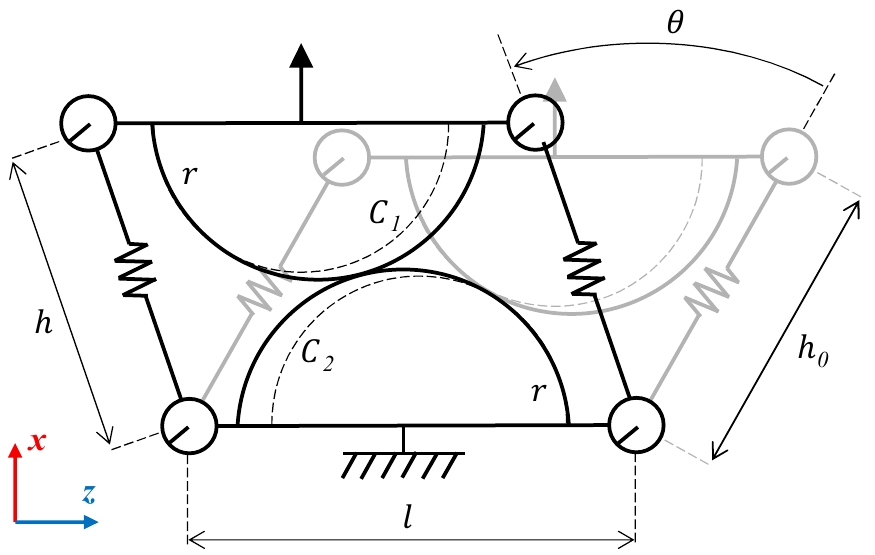}
    \caption{Elasto-mechanical scheme of the absorption unit with rotational spring at the joints angle $\theta$ in black, and $\theta = 0$ in gray.}
    \label{fig:absorption-unit}
    \vspace{-0.3cm}
\end{figure}

The four-bar mechanism gives the inherent wanted spring effect with the compliant hinges. The following presents our proposed strategy for adding the damping through the motion. For a given parallelogram, the rotation over the crank shows some points keeping equidistance. These points describe ideal arcs of circles. We draw two of them with ${\cal{C}}_1$ and ${\cal{C}}_2$ in dotted lines in Fig.~\ref{fig:absorption-unit}. They describe a path with no friction. The core principle is further to propose a bulge shape giving the aimed friction along $\theta$.

Considering the material in its elastic limit, soft hinges give the mechanism both rotational spring and linear spring effect, respectively, around their center of rotation and along links. Firstly, the model is composed of two parallel springs creating a normal force at bulges in a single point. We consider this force parallel to the linear spring. After rotation of an angle $\theta$ in Fig.~\ref{fig:absorption-unit}, elongation of the spring is $\Delta h = | h - h_0 |$. According to Hooke's law, the spring force is
\begin{equation}\label{eq:spring-lin}
    F_k = - k \Delta h = - k r,
\end{equation}
with $k$ the equivalent elastic coefficient of linear springs, and $r \in \nR{+}$ is the positive difference function of $\theta$ defined by the bulge shape. This last is an adjustable parameter depending on the targeted damping. At a given angle $\theta$, a higher value of $r$ will create more friction.

Secondly, the model is composed of four rotational springs creating tangential forces on bulges along the rotation. The torque $\tau_{\kappa}$ is analog to the linear one
\begin{equation}\label{eq:spring-rot}
    \tau_{\kappa} = - \kappa \theta = r F_{\kappa},
\end{equation}
with $\kappa$ the torsion coefficient of the compliant hinge and $F_{\kappa}$ the tangential force at contact of bulges. This torsion force pushes the arm into its idle extended position.

According to Coulomb's law of friction for dry surfaces, the model can be approximated with
\begin{equation}\label{eq:coulomb}
    F_f \leq \mu F_n,
\end{equation}
where $F_f$ is the friction force parallel to the surface and opposed to movement, $\mu$ is the coefficient of friction, and $F_n$ is the normal force applied on each surface. The friction coefficient $\mu$ is empirical and can be found in tables. It depends on the material and the asperities of the two surfaces. The static friction coefficient $\mu_s$ is generally higher than the coefficient of kinetic friction $\mu$. There is a maximum force $F_{max}$ to reach, to begin the slide of one surface onto another,
\begin{equation}\label{eq:static-force}
    F_{max} = \mu_s F_n.
\end{equation}
The surface will start to slide as soon as $F_{max}$ is exceeded. Considering the mechanism in motion, we can then combine~\eqref{eq:spring-lin} and~\eqref{eq:coulomb} such that
\begin{equation}\label{eq:1}
    F_k \leq r \mu k.
\end{equation}

Since the mechanism creates a pure rotation we can consider the total torque $\tau_{total}$ output of the mechanism in motion (i.e., without considering Eq.~\ref{eq:static-force}),
\begin{equation}\label{eq:2}
    \tau_{total} = \tau_{\kappa} + r F_k = r (F_{\kappa} + F_k),
\end{equation}
that is
\begin{equation}\label{eq:3}
    \tau_{total} \leq r^2 \mu k  - r \kappa \theta.
\end{equation}

We can therefore notice that the output torque $\tau_{total}$ depends on the shape of the bulges with $r$ and on the current angle $\theta$.

\subsection{Compliant Mechanism Design}
In the standard mechanism, the use of metallic bolts and bearings is often neglected in the total weight of a mechanism. Nevertheless, for aerial platforms, every gram matters. We want to take advantage of additive manufacturing by combining soft materials and not using any metallic parts. The idea is to reduce the inertial effect of the arm on the platform.

Starting from the kinematic design presented in Fig.~\ref{fig:kinematic-design} and adding the absorption theory of Sec.~\ref{sec:absorption-unit}, we proposed the ideal compliant robot arm of Fig.~\ref{fig:compliant-design}. The mechanism must be fine-tuned, such as in Fig.~\ref{fig:manufactured-design}. Additive manufacturing imperfections on elasto-mechanical behaviors cannot be predicted accurately, so we propose differentiating the real mechanism from the ideal one with several printing iterations. We noticed slight deflection of the \ac{PETG}-carbon fiber links of the four-bar mechanism, as well as small displacement of the \ac{TPU} hinges' center of rotation.

\begin{figure}[h]
    \centering
    \includegraphics[width=\linewidth]{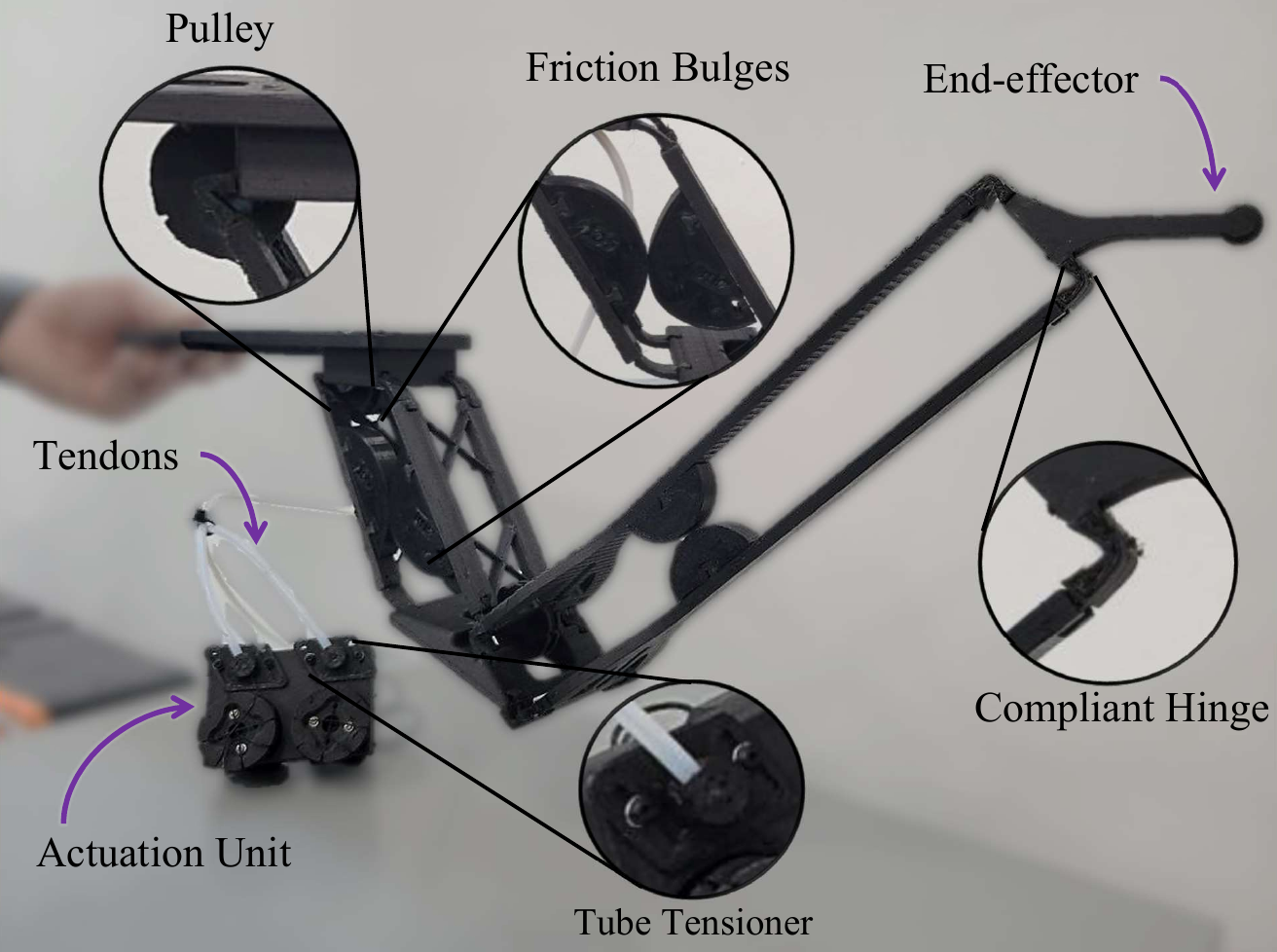}
    \caption{Compliant robot arm mechanical design with main components highlighted.}
    \label{fig:mechanical-design}
    \vspace{-0.2cm}
\end{figure}

Rigid links of the proposed compliant mechanism are 3D-printed in \ac{PETG}-carbon fiber to ensure stiffness, while hinges and bulges are in \ac{TPU} to provide flexibility and friction. All components are tightly assembled.

\subsection{Actuation Unit}
Even if the compliance could have been controlled with a motor-actuated admittance filter, here, the external absorption capabilities are passive. Nevertheless, arm actuation is done in compression, in the opposite direction of idle extension position. This proposed way to actuate the arm firstly makes the position control possible and, secondly, keeps safe passive impact dissipation possible with no effect on the actuators.

As shown in Fig.~\ref{fig:mechanical-design}, to keep the robot arm inertia as low as possible, we deported the actuation unit in the opposite direction. The force is transferred through a tendon mechanism. It comprises pulleys attached by a cable going inside a tube from the actuator to the actuated part. The tube acts as the cable housing, firmly attached from both ends, providing protection and guidance while ensuring that the actuation operates effectively. By adjusting the tension of the inner cable, the actuated pulleys move as the actuators do. Thanks to the rotational spring of the soft joints, motors have only to pull in one direction to reach the equilibrium position.

One can notice that cable housing is fixed such that at actuation, compliance of the mechanism disengages the frictional bulges. Also, to adjust the cable tension at the idle position, the position-adjustable tube housing can be slid. It is located on the actuation unit, tangential to the pulley, and is firmly secured.

\section{EXPERIMENTAL DEMONSTRATION}\label{sec:experiment}
This section presents the evaluation of the proposed flexible solution in the context of \ac{APhI}. In particular, we focus on evaluating two critical aspects: the influence of the robotic arm on the multirotor performance in free flight and the shock absorption performance of the arm when going in contact with a surface. Table~\ref{tab:arm_components} presents the components used for the experimental application.


\begin{table}[ht]
    \centering
    \begin{tabular}{|l|l|l|}
    \hline
    \textbf{Component} & \textbf{Name} & \textbf{Weight [g]} \\
    \hline
     Servo & Dynamixel XC430-W150-T & 231.8 \\
     Robotic Arm & Flexible Arm & 167.6 \\
     Aerial Platform & Tarot 650 Custom & 3553.6 \\\hline
    \end{tabular}
    \caption{Flexible arm components and weights}
    \vspace{-0.6cm}
\label{tab:arm_components}
\end{table}

\subsection{Influence on The Platform}

\begin{figure}
    \captionsetup[subfigure]{aboveskip=-1pt,belowskip=-1pt}
    \centering
    \subfloat[Body frame accelerations on their axis\label{fig:free_flight_test_acc}]{%
        \includegraphics[width=\linewidth]{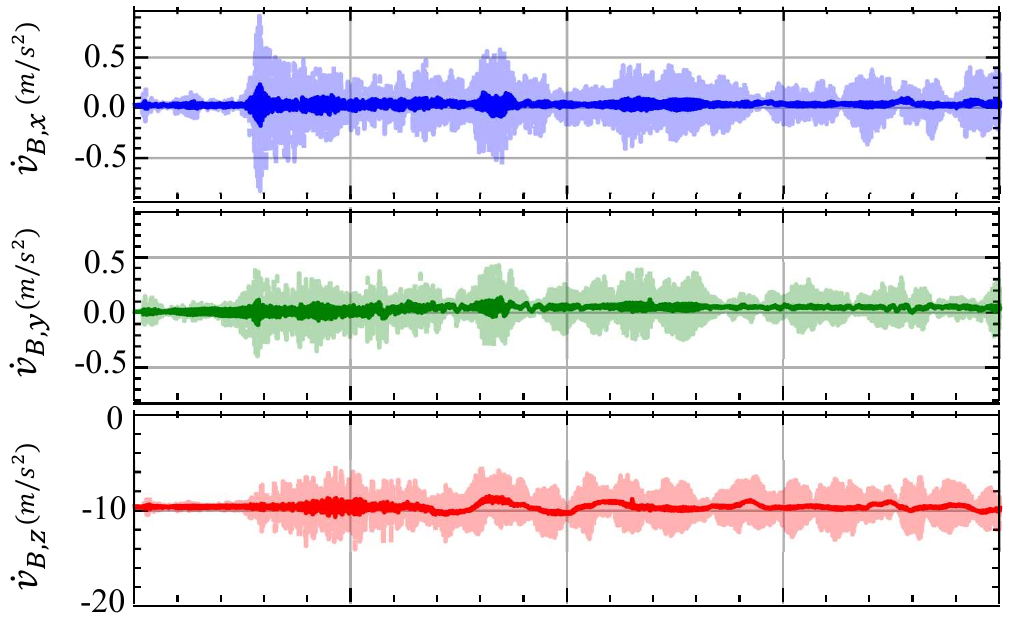}}
    \vfill
    \subfloat[Body frame angular rate on their axis\label{fig:free_flight_test_gyro}]{%
        \includegraphics[width=\linewidth]{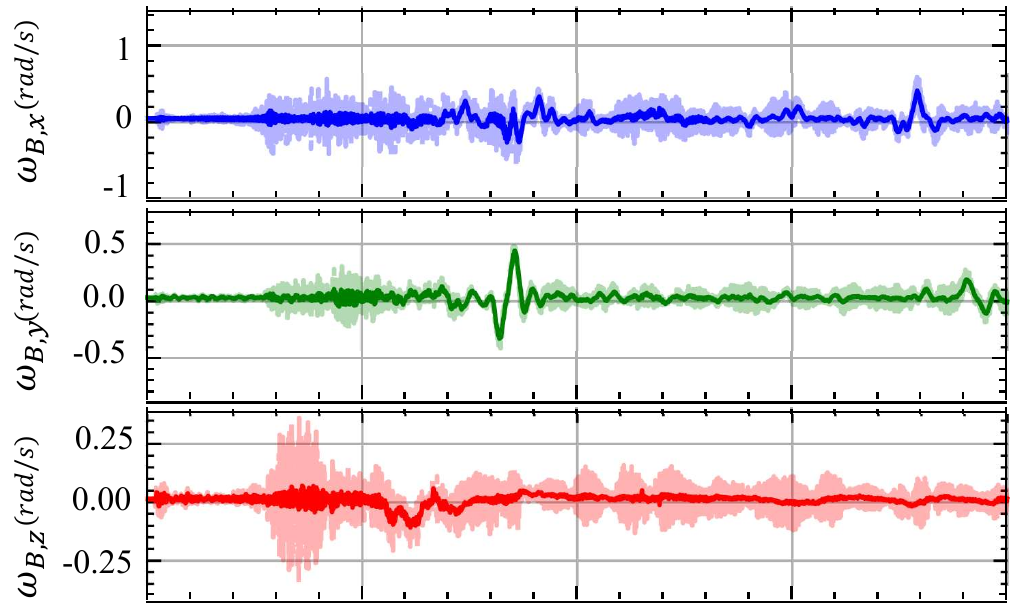}}
    \vfill
    \subfloat[Servo Angles\label{fig:free_flight_test_servo}]{%
        \includegraphics[width=\linewidth]{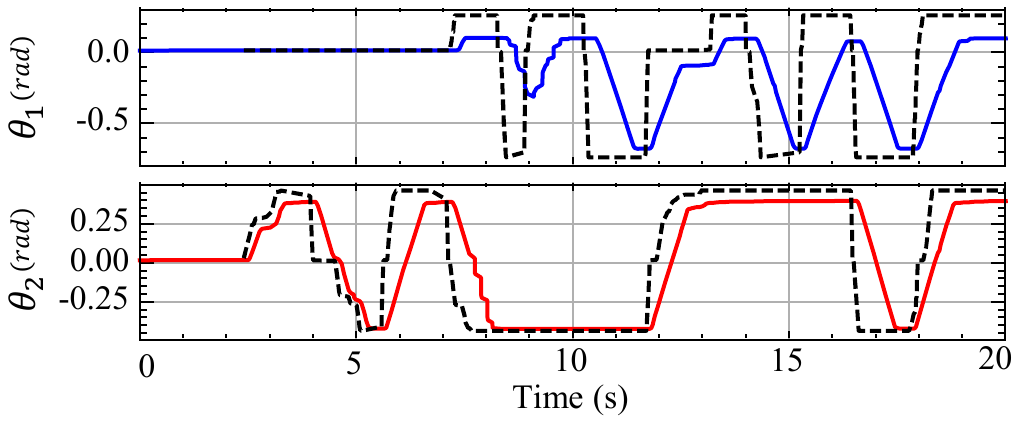}}

    \caption{Flight test results with the proposed soft arm. Acceleration (\ref{fig:free_flight_test_acc}), angular velocity (\ref{fig:free_flight_test_gyro}), servos angular position (\ref{fig:free_flight_test_servo}) and its reference (dashed lines) obtained during flight test. The transparent shadow represents the unfiltered measurements. The solid line is a filtered version for the graph’s clarity}
    \label{fig:free_flight_test}
    \vspace{-0.5cm}
\end{figure}

This first experiment evaluate the physical impact of the robotic arm on the multirotor platform. To test it, we perform a stabilized flight with the \ac{MAV} while an operator controls the robot arm. In this context, even minimized by our design, the motion of the arm generates some inertial forces and torques into the body frame that may disturb the platform's state. The results have been presented in Fig.~\ref{fig:free_flight_test}.

The measured acceleration and angular rate on the \ac{MAV} body frame are plotted in Fig.~\ref{fig:free_flight_test_acc} and Fig.~\ref{fig:free_flight_test_gyro}, while the servo motor actuation is presented in Fig.~\ref{fig:free_flight_test_servo}. Some perturbations have been noticed on the platform. During the experiment, it appears that $\accBx$ and $\accBy$ had similar behavior, remaining centered in $0$ without exceeding $|\acc| = 0.2$~m/s$^2$. A maximum acceleration of $-8$~m/s$^2$ along the z-axis has been measured at $t=8.1$~s.

In the same way, for angular rate measurement, we observe analogous comportment on the x and the y-axis. Both are centered in $0$ and maximum value of $0.49$~rad/s are visible at $t=17.9$~s for $\omega_{B,x}$, and at $t=8.7$~s for $\omega_{B,y}$.

However, they appear to be sporadic behavior, and they do not manifest any trend or correlation with the motion of the arm from Fig.~\ref{fig:free_flight_test_servo}. Therefore, it can be deduced that the forces and torques generated by the arm affect only slightly the platform, demonstrating a stable behavior during free flight.

One crucial aspect manifested in Fig.~\ref{fig:free_flight_test_servo} is the discrepancy between the desired position (dotted line) and the servo-achieved position (solid line) for specific extreme values. This difference is evident at angles corresponding to the points where the total torque $\tau_{total}$ (refer to Eq. ~\eqref{eq:3}) reaches its maximum value during the rotation. This phenomenon indicates a limitation in reaching these angles due to the limited power of servomotors.

\subsection{Shock Absorption}
\begin{figure}
    \captionsetup[subfigure]{aboveskip=-1pt,belowskip=-1pt}
    \centering
    \subfloat[Impact test with rigid arm\label{fig:impact_flight_test_rigid}]{%
        \includegraphics[width=\linewidth]{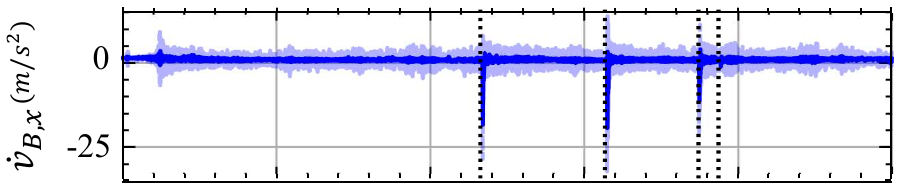}}
    \vfill
    \subfloat[Impact test with flexible arm\label{fig:impact_flight_test_soft}]{%
        \includegraphics[width=\linewidth]{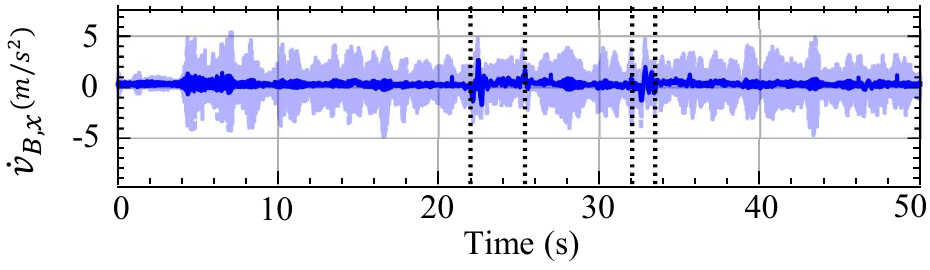}}
    \caption{Comparison of acceleration force generated during impact test with a rigid stick (\ref{fig:impact_flight_test_rigid}) arm and with flexible arm (\ref{fig:impact_flight_test_soft}). Dashed vertical lines highlight the time instances where the platform collides with the interaction surface. The transparent shadow represents the unfiltered measurements while the solid line is a filtered version.}
    \label{fig:impact_test}
    \vspace{-0.5cm}
\end{figure}

To study the energy dissipation capabilities of the proposed platform, we perform impact tests. It consisted of coming in contact with a surface at a given velocity along the x-axis with the aerial manipulator near hovering.
The test has been performed at different velocities, with the proposed flexible arm and a rigid stick for comparison. As depicted in Fig.~\ref{fig:impact_test}, the platform equipped with the rigid arm displays prominent peaks in acceleration upon contact reaching $\acc = -30$~m/s$^2$. In contrast, the platform featuring the proposed solution with a soft arm reveals a lack of acceleration peaks during contact. This test demonstrates better shock absorption capabilities of the flexible arm with respect to the standard solution, resulting in safer interaction of the aerial manipulator with the environment.

By measuring platform velocity before impact and after impact in case of bounce, we can estimate the energy absorbed in impact. We compare two impacts with similar velocities. For the rigid stick third measured interaction, the aerial robot impacted the surface with an absolute velocity of $v_x = 0.626$~m/s and bounced with $v_x = 0.219$~m/s. We noticed a slight deformation of the surface, that dissipated some of the energy. Thus, $E_c = 0.6797$~J have been dissipated. On the other hand, the aerial platform with our compliant mechanism at the fourth measured interaction had a slightly higher velocity $v_x = 0.675$~m/s. The platform remained in contact, so the arm absorbed $E_c = 0.9005$~J. Our system shows better energy dissipation at higher speeds and, therefore, greater stability on impact for safer contact.

\section{CONCLUSION}
We developed a lightweight, shock-absorbing, and actuated compliant robot arm. Our approach for \ac{APhI} includes admittance control for the platform, robot arm inverse kinematics, and careful material selection for weight reduction.

Our work demonstrated that the compliant robot arm has negligible influence on free flight, thanks to tendon-based actuation compensating for arm weight. We explored the arm's compliance during \ac{APhI}, emphasizing the advantages of flexible manipulators with compliant hinges and frictional elements to minimize system weight and complexity while improving impact energy dissipation.

Our future works aim to address end-effector position tracking challenges, closing the control loop to enhance overall accuracy with the platform controller. We also plan to study scalability into other application scenarios. 


\addtolength{\textheight}{-5cm}   






\bibliographystyle{IEEEtran}
\bibliography{bibliography.bib}

\end{document}